\documentclass[pdflatex,sn-nature]{sn-jnl}

\usepackage{graphicx}%
\usepackage{multirow}%
\usepackage{amsmath,amssymb,amsfonts}%
\usepackage{amsthm}%
\usepackage{mathrsfs}%
\usepackage[title]{appendix}%
\usepackage{xcolor}%
\usepackage{textcomp}%
\usepackage{manyfoot}%
\usepackage{booktabs}%
\usepackage{algorithm}%
\usepackage{algorithmicx}%
\usepackage{algpseudocode}%
\usepackage{listings}%
\usepackage{comment}
\usepackage{titletoc}

\usepackage{xcolor}
\definecolor{deepblue}{RGB}{25,25,225}
\newenvironment{blueblock}
{\color{deepblue}}
{}

\usepackage{graphicx}%
\usepackage{multirow}%
\usepackage{amsmath,amssymb,amsfonts}%
\usepackage{amsthm}%
\usepackage{mathrsfs}%
\usepackage[title]{appendix}%
\usepackage{xcolor}%
\usepackage{textcomp}%
\usepackage{manyfoot}%
\usepackage{booktabs}%
\usepackage{algorithm}%
\usepackage{algorithmicx}%
\usepackage{algpseudocode}%
\usepackage{listings}%
\usepackage[most]{tcolorbox}
\tcbuselibrary{listings,breakable}
\newtcblisting{promptbox}{
  breakable,
  enhanced,
  colback=gray!6,
  colframe=gray!45,
  boxrule=0.4pt,
  arc=1mm,
  left=1mm,
  right=1mm,
  top=1mm,
  bottom=1mm,
  listing only,
  listing options={
    basicstyle=\ttfamily\scriptsize,
    breaklines=true,
    breakatwhitespace=false,
    columns=fullflexible,
    keepspaces=true,
    showstringspaces=false,
    aboveskip=0pt,
    belowskip=0pt
  }
}

\theoremstyle{thmstyleone}%
\theoremstyle{thmstyletwo}%

\theoremstyle{thmstylethree}%

\begin{document}

\title[AI Research Agents Narrow Scientific Exploration]{AI Research Agents Narrow Scientific Exploration}

\author[1]{\fnm{Yixuan} \sur{Tang}}\email{ytangch@connect.ust.hk}
\author*[1]{\fnm{Yi} \sur{Yang}}\email{imyiyang@ust.hk}

\affil*[1]{\orgdiv{ISOM}, \orgname{The Hong Kong University of Science and Technology}}


\abstract{AI research agents now support large-scale AI-assisted scientific discovery. We examine whether AI-generated ideas broaden scientific exploration or primarily reinforce existing work. Using five agent frameworks and five large language models, we generate 219,655 ideas for different scientific fields. Across experiments, four consistent patterns emerge. First, AI-generated ideas are more concentrated than human-authored papers within the same research area. Second, they remain much closer to starting literature than later human follow-on work does. Third, AI-generated ideas align less with future human research. Last, AI-generated ideas are located in lower-impact regions of the historical scientific landscape. Overall, current AI research agents appear better suited to local elaboration than to broadening scientific exploration. 
}




\maketitle

\section{Introduction}\label{sec1}

\begin{figure}[t]
\centering
 \includegraphics[width=\linewidth]{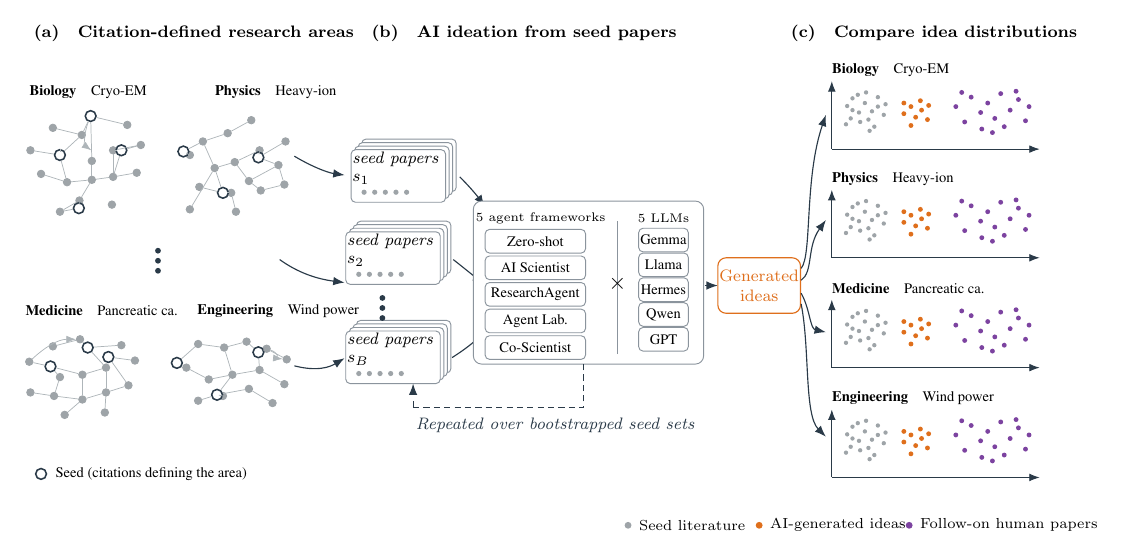}
\vspace{-0.5cm}
\caption{Overview of the study design. We first construct research areas from the Semantic Scholar corpus, use AI research agents to generate scientific ideas from prior literature in those areas, and compare the resulting ideas against human-authored papers.}
\label{fig:study_design}
\end{figure}

Recent advances in AI research agents have raised the possibility of automating scientific discovery. These agents can now conduct literature reviews, generate research ideas, plan experiments, run code, write papers, and iteratively explore and refine scientific hypotheses~\cite{lu2026towards,baek2025researchagent,schmidgall2025agent, hao2026artificial}.  
Importantly, these AI research agent frameworks are explicitly designed to encourage exploratory scientific ideation. Their prompts and reasoning procedures often instruct agents to generate novel, high-impact, and unconventional ideas rather than simple extensions of prior work~\cite{lu2026towards,baek2025researchagent,schmidgall2025agent}.
As such systems become increasingly capable and accessible, they may fundamentally reshape the process of scientific discovery.

Yet the ability to generate research ideas at scale does not necessarily imply broader scientific exploration. Scientific discovery often depends on moving beyond established directions, searching less familiar regions, and recombining prior knowledge in non-routine ways~\cite{foster2015tradition,uzzi2013atypical}. Existing evaluations of AI research agents mainly assess whether individual ideas are interesting, novel, feasible, or executable~\cite{si2024canllms,wang2024scimon}, but reveal much less about how repeated AI-assisted ideation may shape the broader landscape of scientific exploration. 
This raises a broader question: do AI research agents broaden scientific exploration?

To study this question, we construct research areas from the scientific literature across broad fields and use AI research agents to generate ideas from shared seed papers within each research area. Specifically, using papers published between 2020 and 2025 from the Semantic Scholar Academic Graph as seed literature, we use advanced AI research-agent frameworks, including AIScientist~\cite{lu2026towards}, ResearchAgent~\cite{baek2025researchagent}, AgentLaboratory~\cite{schmidgall2025agent}, and Co-Scientist~\cite{gottweis2026coscientist}, together with five LLMs to generate complete scientific research ideas, including both research questions and methods. In total, we analyze 219,655 valid AI-generated research ideas spanning 12 broad scientific fields and 155 research areas. Throughout the ideation process, all evaluated AI agent frameworks are explicitly instructed to explore novel research directions beyond the seed literature. The AI agent frameworks can further search and retrieve additional relevant literature from the entire Semantic Scholar database.

We then investigate these AI-generated ideas from four perspectives: whether they explore diverse scientific directions, whether they move beyond their starting literature toward new research topics, whether they align with future research frontiers, and whether they are associated with potentially high-impact regions of the scientific landscape.

Across research fields and consistently across evaluated agent frameworks and underlying LLMs, four patterns emerge. First, AI-generated ideas are substantially more concentrated than human-authored papers from the same research areas. Second, AI-generated ideas remain much closer to their starting literature than later human follow-on work does, indicating that they primarily extrapolate locally from prior work. Third, AI-generated ideas cover substantially fewer keywords characterizing the next year’s research frontier, defined by the most frequently studied topics in subsequent human research, than follow-on human papers. Finally, AI-generated ideas are associated with human-authored papers that receive fewer citations than human follow-on work.


These findings suggest a cautious view of the role of current AI research agents in scientific discovery. Although current AI agents can efficiently generate coherent, literature-grounded research ideas at scale, their outputs do not appear to substantially expand the frontier of scientific exploration. As AI systems become increasingly integrated into scientific workflows, the distinction between scaling idea generation and broadening scientific exploration may become increasingly important. The broader challenge is therefore not simply to make AI systems generate more plausible scientific ideas, but to design systems that genuinely expand the range of scientific exploration.

\begin{table}[t]
\centering
\caption{\textbf{AI research agent frameworks evaluated in this study.}}
\label{tab:agent_scaffolds}
\footnotesize
\setlength{\tabcolsep}{4pt}
\renewcommand{\arraystretch}{1.15}

\begin{tabular}{p{2.2cm} p{3cm} p{4cm} p{4cm}}
\toprule
\textbf{Framework} & \textbf{Agentic mechanism} & \textbf{Implementation summary} & \textbf{Novelty instruction excerpt} \\
\midrule

Zero-shot 
& Single-pass generation 
& The LLM receives literature context and generates a research idea in a single interaction. 
& ``propose one novel research idea grounded in the literature'' \\

AI Scientist~\cite{lu2026towards} 
& Iterative self-reflection 
& The agent iteratively critiques and revises generated ideas, optionally refreshing literature context between rounds. 
& ``propose any novel ideas or experiments; make sure they are novel''; ``quality, novelty, and feasibility'' \\

ResearchAgent~\cite{baek2025researchagent} 
& Multi-stage planning and validation 
& The agent decomposes ideation into problem finding, method design and experiment planning, with intermediate validation agents scoring outputs. 
& ``promising, new, and key scientific problems''; ``original''; ``innovative''; validation dimensions include ``Originality'' and ``Innovativeness'' \\

AgentLaboratory~\cite{schmidgall2025agent} 
& Multi-agent deliberation 
& Multiple role-based agents iteratively discuss and refine research proposals through dialogue. 
& ``very innovative and unlike anything seen before''; ``Make sure your new output is very different'' \\

Co-Scientist~\cite{gottweis2026coscientist}
&Tournament-based hypothesis evolution
& Multiple agents generate competing hypotheses, then review, rank, and evolve them through tournament comparison before a final meta-review.
& ``develop one novel, feasible research hypothesis'' and ``rank hypotheses by novelty, significance, feasibility, and testability'' \\

\bottomrule
\end{tabular}

\begin{flushleft}
\footnotesize
The final column presents excerpts of explicit novelty-related instructions from each framework.
\end{flushleft}
\end{table}

\section{Generating Scientific Ideas with AI Agents}

\paragraph{Define Research Areas.}
We begin by constructing research areas from the scientific literature across major fields of science. We collect papers from the Semantic Scholar Academic Graph\footnote{https://www.semanticscholar.org/product/api}, together with their reference and citation information. The resulting corpus spans 12 fields, including Medicine, Biology, Engineering, Chemistry, Computer Science, Environmental Science, Materials Science, Physics, Mathematics, Economics, Business, and Sociology. Each paper record includes its abstract, publication year, primary field, and citation links. We use papers published before 2020 to construct research areas.

Within each scientific field, we identify research areas using bibliographic coupling~\citep{kessler1963bibliographic}. 
Specifically, papers are represented by their citation profiles and clustered according to bibliographic-coupling similarity. The final identified research areas span topics including cryo-electron microscopy, pancreatic cancer treatment, offshore wind power, heavy-ion physics, and microbiome community assembly. Detailed construction procedures are provided in the Supplementary Methods {(see SM S1.1)}.

\paragraph{Scientific Idea generation.}

We next use AI research agents to generate new scientific ideas from prior literature. For each identified research area, 
We repeatedly sample seed-paper sets from papers published between 2020 and 2025 to initialize AI ideation. Each seed set contains five papers: one anchor paper together with four related papers from the same research area, selected using citation. We use five seed papers because most evaluated AI research agent frameworks are constrained by context-window limitations of current LLMs. These seed papers define a coherent research topic and provide a common starting literature context for idea generation.

During idea generation, the evaluated agent frameworks can further search a locally deployed Semantic Scholar database to retrieve additional relevant papers, allowing them to expand the literature context beyond the initial seed set. To preserve the historical setting, agents are only allowed to retrieve papers that were available when the seed papers were published.

We evaluate five representative AI research-agent frameworks: a Zero-shot baseline, AIScientist~\cite{lu2026towards}, ResearchAgent~\cite{baek2025researchagent}, AgentLaboratory~\cite{schmidgall2025agent}, and Co-Scientist~\cite{gottweis2026coscientist}. These frameworks represent several major designs for AI research agents, including iterative self-reflection, multi-stage planning and validation, multi-agent deliberation, and tournament-based hypothesis evolution. All evaluated agent frameworks, except the Zero-shot baseline, can retrieve additional papers from the locally deployed Semantic Scholar database. Table~\ref{tab:agent_scaffolds} summarizes the evaluated frameworks and their corresponding agentic mechanisms.

Importantly, across all evaluated frameworks, the ideation prompts explicitly encourage exploration beyond the seed literature. The Zero-shot baseline asks the model to propose a novel research idea. AIScientist emphasizes generating ``novel'' and ``high-impact'' ideas through iterative self-reflection and revision. ResearchAgent encourages innovative method design during multi-stage planning and validation. Agent Laboratory explicitly instructs agents to expand upon the literature and generate ideas that are ``very innovative and unlike anything seen before.'' Co-Scientist generates and refines hypotheses through comparison, and explicitly rewards novelty at each round. Supplementary Methods provides the full prompts and detailed agentic design of each evaluated framework {(see SM S1.2)}.

Each AI agent framework needs to be paired with an LLM. We evaluate five LLMs: four open-weight models ranging from 8B to 35B parameters, Gemma-4-31B-IT~\cite{farabet_lacombe_gemma4_2026}, Llama-3.1-8B~\cite{grattafiori2024llama}, Hermes-4-14B~\cite{teknium2025hermes}, and Qwen3-35B-A3B~\cite{qwen3.5}, together with OpenAI's GPT-5.4~\cite{openai_chat}.

Across the five agent frameworks and five LLMs, we bootstrap seed-paper sets from the identified research areas. In total, our analysis uses 219,655 valid AI-generated ideas generated from 155  research areas spanning 12 broad scientific fields, obtained from  232,800 generation runs. A generation run is considered valid if it successfully produces a non-empty structured research idea{ (see SM S1.2, tables S3 and S4 for the detailed breakdown and examples)}. 

\begin{figure}[t]
    \centering
    \includegraphics[width=.9\linewidth]{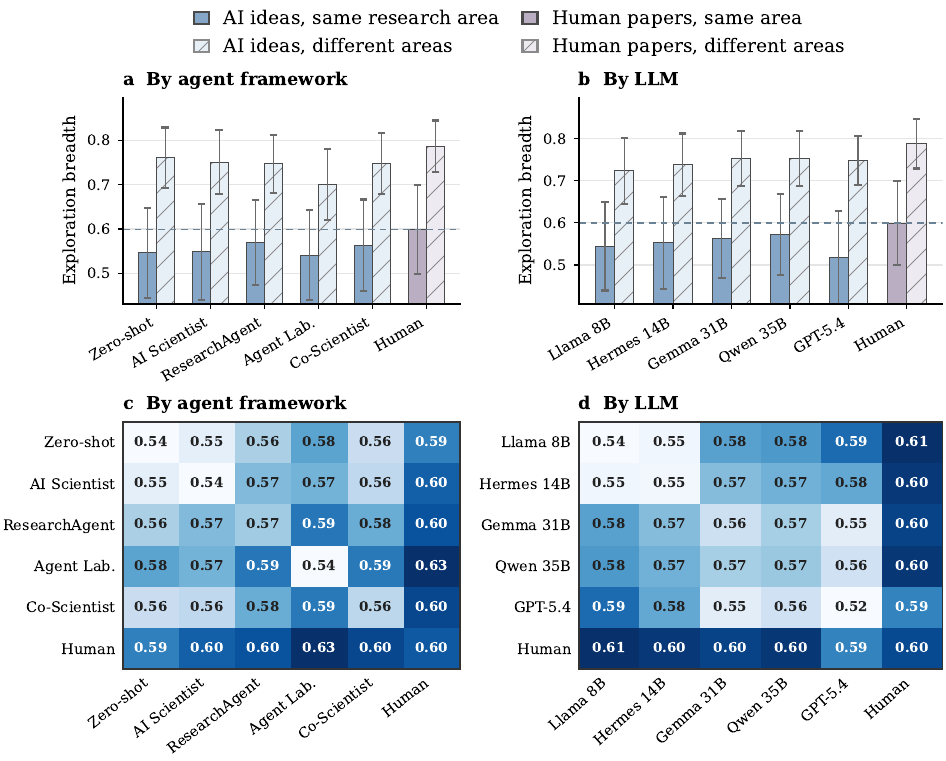}
    \caption{AI-generated ideas exhibit lower exploration breadth than human-authored papers. (a–b) Average exploration breadth within the same research area and across different research areas, shown by agent framework and by LLM, with the corresponding human-paper baseline. (c–d) Pairwise exploration-breadth matrices between ideas generated by different agent frameworks and different LLMs. The last row and column compare AI-generated ideas with human-authored papers.}
    \label{fig:fig_idea_concentration}
\end{figure}

\section{Quantifying AI-Generated Scientific Ideas}
We next introduce several measurements to characterize AI-generated scientific ideas.

\textbf{Exploration breadth.}
We measure exploration breadth as the extent to which AI-generated ideas spread across distinct directions within an identified research area~\citep{peng2021neural}. To quantify this breadth, we encode every AI-generated idea into a shared semantic embedding space using a text embedding model. Exploration breadth is then measured as the average pairwise cosine distance among AI-generated ideas within the same research area. Higher average cosine distance indicates that generated ideas occupy a broader region of the semantic idea space and therefore explore a wider range of scientific directions~\citep{pnasHofstra2020,cohan2020specter}. As a robustness check, we also quantify exploration breadth using a centroid-based distance measure {(see SM S2.2 and table S6)}.

\textbf{Exploration distance.}
We measure exploration distance as the extent to which  AI-generated ideas move beyond the seed literature used to initialize ideation~\citep{shibayama2021measuring}. For each idea, we first compute the centroid of the five human-authored seed papers in the semantic embedding space. Exploration distance is then quantified as the cosine distance between each AI-generated idea and the corresponding seed-paper centroid. Larger distances indicate that generated ideas move further away from their starting literature.

\textbf{Frontier alignment.}
We measure frontier alignment as the extent to which AI-generated ideas align with emerging research directions. For each broad scientific field, we define the next-year field frontier as the set of the top 10\% most frequent scholarly keywords extracted from human-authored papers published in the year following the seed-paper set~\citep{cui2026aging}. To obtain a common representation, we use an LLM to extract scholarly keywords from both AI-generated ideas and human-authored papers. For each identified research area, we aggregate the extracted keywords from all AI-generated ideas into a single AI keyword set. Frontier alignment is then computed as the proportion of frontier keywords that also appear in the aggregated AI keyword set. Higher frontier alignment indicates that AI-generated ideas align more closely with the future human research frontier.

\textbf{Potential scientific impact.}
We measure the potential scientific impact of AI-generated ideas based on the citation performance of semantically similar human-authored papers~\citep{arts2025beyond}. Because AI-generated ideas themselves have no citation records, we use nearby human-authored papers in the semantic embedding space as observable proxies. We first compute a normalized citation score for each human-authored paper relative to papers published in the same research area and publication year. We then identify the 20 nearest human-authored papers for each AI-generated idea and define its potential impact score as the average normalized citation score of these neighbors. Higher impact scores indicate that AI-generated ideas are located in potentially higher-impact regions of the scientific landscape.

\section{Empirical Analysis}

\subsection{Exploration Breadth: AI Ideas Are More  Concentrated Than Human Papers}
\label{sec:concentrate}

\begin{tableorg}[t]
\centering
\caption{\textbf{Exploration breadth by agent framework and LLM.} Exploration breadth is first computed within each research area and then averaged across research areas. The AI and Human columns report the mean exploration breadth of AI-generated ideas and the corresponding human-authored papers, respectively. AI--Human denotes the difference between  AI-generated ideas and human-authored papers. Brackets report 95\% bootstrap confidence intervals. }
\label{tab:exploration_breadth_summary}
\small
\begin{tabular*}{.6\textwidth}{@{\extracolsep{\fill}}lrrr@{}}
\toprule
Group &  AI & Human & AI--Human [95\% CI] \\
\midrule
\multicolumn{4}{l}{\textit{Pooled}} \\
All data &  0.554 & 0.599 & -0.045 [-0.049, -0.042] \\
\midrule
\multicolumn{4}{l}{\textit{By agent framework}} \\
Agent Laboratory &  0.541 & 0.599 & -0.058 [-0.063, -0.053] \\
AIScientist  &  0.548 & 0.599 & -0.051 [-0.055, -0.046] \\
Co-Scientist &  0.563 & 0.599 & -0.036 [-0.040, -0.030] \\
ResearchAgent &  0.570 & 0.599 & -0.029 [-0.033, -0.025] \\
Zero-shot &  0.546 & 0.599 & -0.053 [-0.058, -0.048] \\
\midrule
\multicolumn{4}{l}{\textit{By LLM}} \\
GPT-5.4 &  0.518 & 0.585 & -0.067 [-0.080, -0.060] \\
Gemma-4-31B-IT &  0.562 & 0.599 & -0.037 [-0.041, -0.032] \\
Hermes-4-14B &  0.553 & 0.599 & -0.047 [-0.051, -0.041] \\
Llama-3.1-8B &  0.545 & 0.599 & -0.054 [-0.059, -0.050] \\
Qwen3-35B-A3B & 0.572 & 0.599 & -0.027 [-0.032, -0.023] \\
\bottomrule
\end{tabular*}
\end{tableorg}

We examine the exploration breadth of AI-generated ideas within each identified research area. As a comparison, we also measure the exploration breadth of the human-authored seed papers. To ensure a fair comparison, we randomly sample one seed paper for each AI-generated idea so that the human and AI collections contain the same number of ideas. 

Figure~\ref{fig:fig_idea_concentration}a--b shows a consistent pattern across the five agent frameworks and five LLMs. AI-generated ideas within the same research area are more similar to one another than human-authored papers from those same areas. Averaged across agent frameworks, the breadth is 0.554 for AI-generated ideas and 0.599 for human-authored papers, representing a 7.5\% lower exploration breadth for AI-generated ideas. 


Panels~c--d further show that different LLMs and agent frameworks often explore overlapping regions of the same research area. Within the same research area, the average exploration breadth between ideas generated by different agent frameworks is 0.572, while that between ideas generated by different LLMs is 0.570.
These averages remain {lower} than the human same-area baseline. Field-level analyses show the same pattern across 11 of the 12 broad scientific fields, with Mathematics being the only exception {(see SM S2.1 and table S5)}. 

We observe the same pattern using an alternative centroid-based measure of exploration breadth. For each research area, AI-generated ideas lie closer to their area centroids than human-authored papers do, again indicating a more concentrated exploration pattern {(see SM S2.2 and table S6)}. 

\subsection{Exploration Distance: AI Ideas Stay Close to Their Starting Literature}


We next examine the exploration distance of AI-generated ideas to assess whether they move beyond the seed literature used to initialize ideation or instead remain locally anchored to it. As a comparison, we examine follow-on human-authored papers that directly cite at least one of the seed papers~\cite{wu2019large}. These follow-on papers represent subsequent human research emerging from the same research topic.

\begin{figure}
\centering
\includegraphics[width=.9\linewidth]{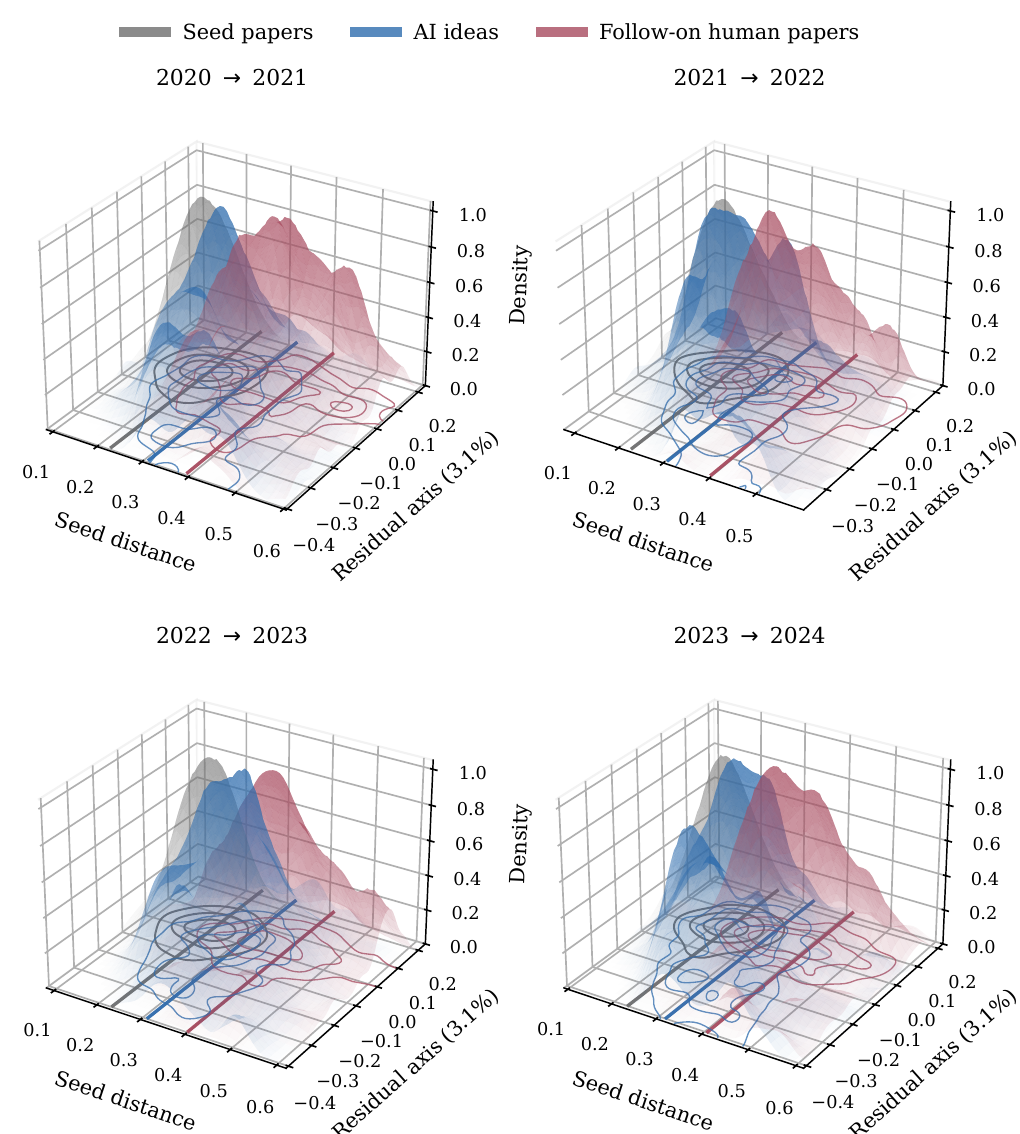}
\caption{AI-generated ideas remain closer to the seed literature than follow-on human research. Distributions of exploration distance for AI-generated ideas (blue) and follow-on human papers (red) across four consecutive year pairs. The horizontal axis shows exploration distance, while the second horizontal axis shows the first residual principal component used only for visualization. Vertical colored lines denote group means.} 
\label{fig:yearly_ai_human_surfaces}
\end{figure}

Figure~\ref{fig:yearly_ai_human_surfaces} compares the distributions of exploration distance for AI-generated ideas and follow-on human papers across four consecutive years  (2020$\rightarrow$2021 through 2023$\rightarrow$2024). Across all four years,  AI-generated ideas remain closer to the seed literature than subsequent human research. In every year, the AI distributions are shifted toward smaller exploration distances, whereas follow-on human papers exhibit broader distributions extending to substantially larger distances. On average, across years, exploration distance is 0.322 for AI-generated ideas, compared with 0.410 for follow-on human papers. Field-level analyses reveal the same pattern across all twelve broad scientific fields, with the difference remaining statistically significant in every field {(see SM S2.1 and table S5)}. This result suggests that, although all evaluated AI agent frameworks can search for and retrieve relevant literature from the Semantic Scholar database, AI-generated ideas remain largely confined to local exploration.

\begin{tableorg}[t]
\centering
\caption{\textbf{Exploration distance from seed literature by year, agent framework, and LLM.} Exploration distance is first computed for each seed-paper set and then averaged across seed-paper sets. The AI and Human columns report the mean exploration distance of AI-generated ideas and the corresponding follow-on human papers, respectively. AI--Human denotes the difference between AI-generated ideas and follow-on human papers. Brackets report 95\% bootstrap confidence intervals.}
\label{tab:exploration_distance_summary}
\small
\begin{tabular*}{.6\textwidth}{@{\extracolsep{\fill}}lrrr@{}}
\toprule
Group &  AI & Human & AI--Human [95\% CI] \\
\midrule
\multicolumn{4}{l}{\textit{By year}} \\
All years &  0.322 & 0.410 & -0.088 [-0.092, -0.083] \\
2020$\rightarrow$2021 &  0.318 & 0.399 & -0.081 [-0.090, -0.072] \\
2021$\rightarrow$2022 &  0.320 & 0.407 & -0.087 [-0.097, -0.077] \\
2022$\rightarrow$2023 &  0.322 & 0.411 & -0.089 [-0.098, -0.079] \\
2023$\rightarrow$2024 &  0.328 & 0.422 & -0.093 [-0.103, -0.084] \\
\midrule
\multicolumn{4}{l}{\textit{By agent framework}} \\
Agent Laboratory &  0.382 & 0.410 & -0.027 [-0.033, -0.022] \\
AIScientist &  0.320 & 0.410 & -0.090 [-0.095, -0.085] \\
Co-Scientist &  0.328 & 0.410 & -0.082 [-0.087, -0.077] \\
ResearchAgent &  0.313 & 0.410 & -0.096 [-0.101, -0.091] \\
Zero-shot & 0.265 & 0.410 & -0.144 [-0.149, -0.140] \\
\midrule
\multicolumn{4}{l}{\textit{By LLM}} \\
GPT-5.4 & 0.289 & 0.411 & -0.121 [-0.142, -0.100] \\
Gemma-4-31B-IT &  0.314 & 0.410 & -0.095 [-0.100, -0.091] \\
Hermes-4-14B &  0.318 & 0.410 & -0.092 [-0.097, -0.087] \\
Llama-3.1-8B &  0.337 & 0.410 & -0.072 [-0.078, -0.067] \\
Qwen3-35B-A3B & 0.319 & 0.410 & -0.091 [-0.096, -0.086] \\
\bottomrule
\end{tabular*}
\end{tableorg}

\subsection{Frontier Alignment: AI Ideas Are Less Aligned with Future Research Frontiers}

\begin{tableorg}[htbp]
\centering
\caption{\textbf{Next-year field frontier alignment by agent framework and LLM.} Frontier alignment is first computed for each broad scientific field and then averaged across fields. The AI and Human columns report the mean frontier coverage of AI-generated ideas and the corresponding follow-on human papers, respectively. AI--Human denotes the difference between AI-generated ideas and follow-on human papers. Brackets report 95\% bootstrap confidence intervals. }
\label{tab:field_frontier_coverage}
\small
\begin{tabular*}{.6\textwidth}{@{\extracolsep{\fill}}lrrr@{}}
\toprule
Group &  AI & Human & AI--Human [95\% CI] \\
\midrule
\multicolumn{4}{l}{\textit{\textbf{Frontier alignment}}} \\
All data & 0.285 & 0.365 & -0.080 [-0.085, -0.075] \\
\midrule
\multicolumn{4}{l}{\textit{By agent framework}} \\
Agent Laboratory &  0.250 & 0.377 & -0.127 [-0.137, -0.117] \\
AI Scientist v2 &  0.328 & 0.360 & -0.032 [-0.041, -0.022] \\
Co-Scientist &  0.282 & 0.375 & -0.093 [-0.104, -0.081] \\
ResearchAgent &  0.234 & 0.336 & -0.102 [-0.111, -0.093] \\
Zero-shot &  0.331 & 0.377 & -0.046 [-0.054, -0.038] \\
\midrule
\multicolumn{4}{l}{\textit{By LLM}} \\
GPT-5.4 &  0.073 & 0.142 & -0.068 [-0.077, -0.060] \\
Gemma-4-31B-IT &  0.259 & 0.391 & -0.131 [-0.140, -0.123] \\
Hermes-4-14B &  0.302 & 0.343 & -0.041 [-0.049, -0.033] \\
Llama-3.1-8B &  0.363 & 0.391 & -0.028 [-0.036, -0.021] \\
Qwen3-35B-A3B &  0.269 & 0.391 & -0.122 [-0.132, -0.113] \\
\bottomrule
\end{tabular*}
\end{tableorg}

We next examine the frontier alignment of AI-generated ideas to assess whether they align with future directions of scientific research. As a comparison, we measure the frontier alignment for follow-on human-authored papers that directly cite at least one of the seed papers. To ensure a fair comparison, these follow-on papers are excluded from the next-year corpus when constructing the frontier keyword set. AI-generated ideas and follow-on human papers contain a comparable number of extracted keywords on average (11.80 versus 11.88).

Across research fields, AI-generated ideas cover 28.5\% of next-year frontier keywords, compared with 36.5\% for follow-on human papers. The difference remains statistically significant across all evaluated research fields. 
One potential concern is information leakage arising from LLM pretraining on papers published after the seed literature. To mitigate this issue, all AI agents are restricted to retrieving papers available at the time of ideation, preventing explicit access to future publications. Although pretrained LLMs may still implicitly encode knowledge from later papers, AI-generated ideas nevertheless exhibit substantially lower frontier alignment than subsequent human research. Therefore, the reported differences are likely conservative estimates.
Together, these results suggest that AI-generated ideas are less aligned with future research directions than the subsequent work produced by human researchers. 


\subsection{Potential Scientific Impact: AI Ideas are Located in Lower-Impact Regions of the Scientific Landscape}
\label{sec:impact}

We next examine the potential scientific impact of AI-generated ideas. As a comparison, we also measure the scientific impact of follow-on human-authored papers that directly cite at least one of the seed papers. For human-authored papers, scientific impact is measured using citation counts normalized by publication year and research field. 
The results show that AI-generated ideas receive lower impact scores than follow-on human papers. Averaged across research areas and study years, the mean potential impact score of AI-generated ideas is 0.387, compared with 0.492 for follow-on human papers, 21.3\% lower than that of follow-on human papers (Table~\ref{tab:local_impact_landscape}). This pattern holds across 11 of the 12 evaluated scientific fields, with Mathematics being the only field in which the difference is not statistically significant.

Moreover, we validate the neighborhood-based impact proxy using a leave-one-out analysis on human-authored papers. Potential scientific impact scores computed from neighboring papers positively predict the target paper's own normalized citation score (Spearman $\rho=0.155$, $p<0.001$. see SM S2.3 and table S7), supporting the validity of using local neighborhoods to estimate the potential impact of AI-generated ideas.

\begin{tableorg}[t]
\centering
\caption{\textbf{Potential scientific impact by agent framework and LLM.} Potential scientific impact is first computed for each research area and publication year and then averaged across research areas and years. The AI and Human columns report the mean potential scientific impact scores of AI-generated ideas and the true normalized citations of corresponding follow-on human papers, respectively. AI--Human denotes the difference between AI-generated ideas and follow-on human papers. Brackets report 95\% bootstrap confidence intervals.}\label{tab:local_impact_landscape}
\small
\begin{tabular*}{.6\textwidth}{@{\extracolsep{\fill}}lrrl@{}}
\toprule
Group & AI  & Human  & AI--Human [95\% CI] \\
\midrule
\multicolumn{4}{l}{\textit{Pooled}} \\
Full Data & 0.387 & 0.492 & -0.105 [-0.114, -0.096] \\
\midrule
\multicolumn{4}{l}{\textit{By agent framework}} \\
Agent Laboratory & 0.411 & 0.492 & -0.081 [-0.090, -0.073] \\
AI Scientist v2 & 0.412 & 0.492 & -0.081 [-0.089, -0.072] \\
Co-Scientist & 0.424 & 0.492 & -0.068 [-0.076, -0.060] \\
ResearchAgent & 0.427 & 0.492 & -0.065 [-0.074, -0.057] \\
Zero-shot & 0.423 & 0.492 & -0.069 [-0.078, -0.061] \\
\midrule
\multicolumn{4}{l}{\textit{By LLM}} \\
GPT-5.4 & 0.472 & 0.464 & 0.008 [-0.029, 0.046] \\
Gemma-4-31B-IT & 0.441 & 0.492 & -0.051 [-0.059, -0.043] \\
Hermes-4-14B & 0.393 & 0.492 & -0.099 [-0.108, -0.091] \\
Llama-3.1-8B & 0.383 & 0.492 & -0.109 [-0.118, -0.101] \\
Qwen3-35B-A3B & 0.450 & 0.492 & -0.043 [-0.051, -0.035] \\
\bottomrule
\end{tabular*}
\end{tableorg}

\subsection{Consistency Across Agent Frameworks and LLMs}
We examine whether the four findings remain consistent across different AI agent frameworks and underlying LLMs (Table~\ref{tab:exploration_breadth_summary}--\ref{tab:local_impact_landscape}). Overall, the qualitative patterns remain remarkably stable. More sophisticated agent frameworks do not substantially reduce the gaps between AI-generated ideas and human-authored research across the four measures, although modest improvements appear in individual dimensions.

For example, the Zero-shot baseline, the only framework without access to the Semantic Scholar search tool, exhibits the largest exploration-distance gap, indicating that its generated ideas remain most closely anchored to the initial seed papers. Agent frameworks that can retrieve additional literature reduce this exploration-distance gap, suggesting that literature search helps agents move beyond the initial seed context. However, this improvement does not translate into substantially higher frontier alignment or potential scientific impact. Thus, providing access to additional literature alone does not fundamentally alter the overall exploration pattern.

A similar pattern is observed across LLMs. GPT-5.4 is the only evaluated model whose generated ideas occupy semantic neighborhoods with potential scientific impact comparable to subsequent human research. Nevertheless, GPT-5.4 explores a narrower region of the scientific idea space and exhibits substantially lower frontier coverage than human follow-on work.

Taken together, these findings suggest one broader implication. Although all evaluated AI agent frameworks are explicitly instructed to generate novel and high-impact ideas, many explicitly reward novelty through iterative self-reflection and refinement, and all agent frameworks can actively search and retrieve relevant prior literature, current AI research agents do not substantially narrow the gap to human researchers in scientific exploration.

\subsection{AI Ideas Primarily Introduce News Methods Rather Than Research Questions}
\label{sec:recombination}

Finally, we examine how AI-generated ideas are constructed. Scientific novelty may arise from identifying new research questions, developing new technical methods, or recombining existing ideas in novel ways~\citep{luo2022combination,uzzi2013atypical}. We therefore prompt an LLM to annotate each AI-generated idea into two components: a research question, describing the scientific problem being studied, and one or more methods, describing how the problem is approached. We then compare the extracted research questions and methods against those appearing in the corresponding five seed papers used during ideation {(see SM S1.2 and S2.4 for details)}.

\begin{figure}[htbp]
\centering
\includegraphics[width=.9\linewidth]{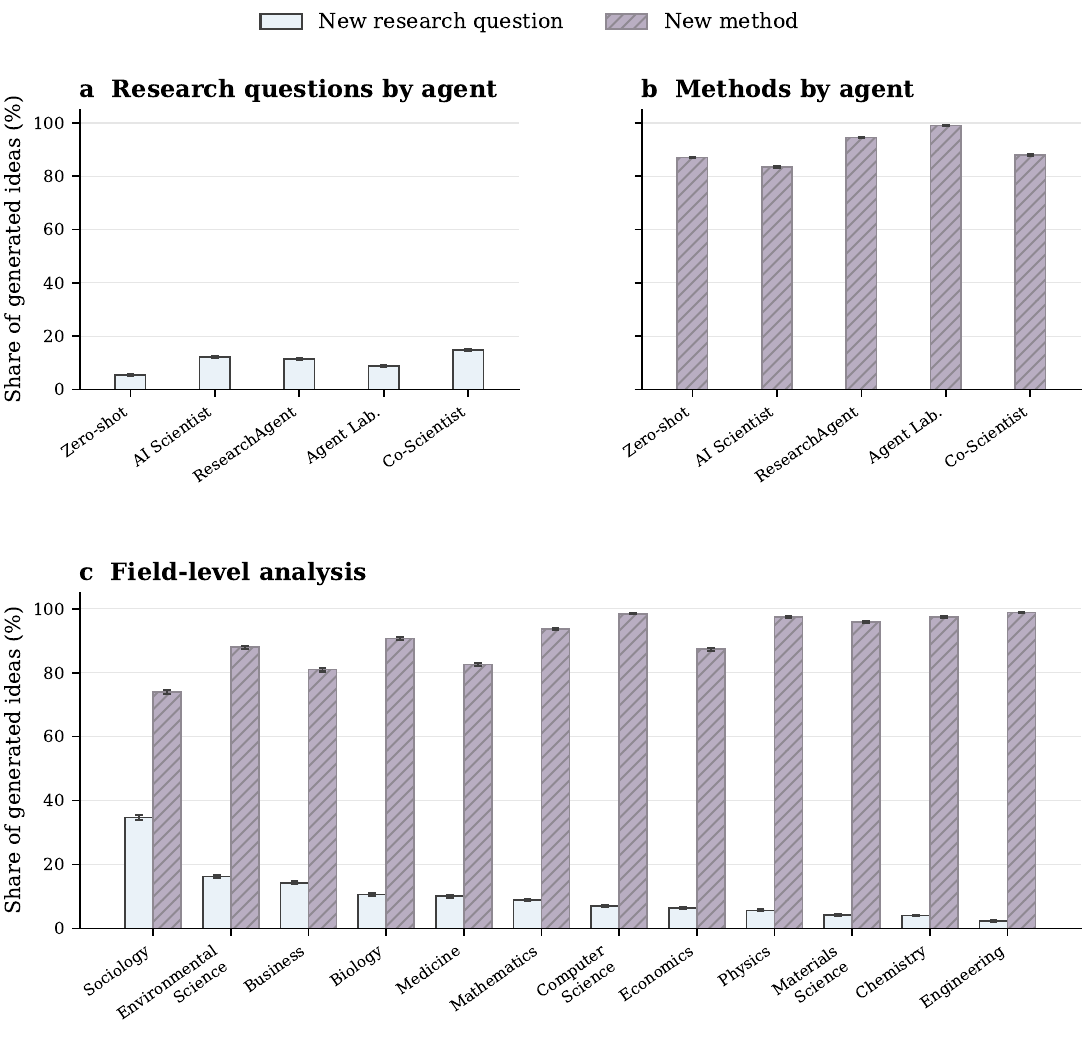}
\caption{\textbf{New research questions and methods introduced by AI-generated ideas relative to the seed literature.}
(a--b) Shares of AI-generated ideas introducing new research questions or methods that do not appear in the corresponding seed literature, shown by agent framework. (c) Field-level shares of AI-generated ideas introducing new research questions and new methods. }
\label{fig:task_method_novelty}
\end{figure}

Overall, AI-generated ideas rarely introduce substantially new research questions. Only 10.5\% of AI-generated ideas contain research questions that are absent from the seed literature (Figure \ref{fig:task_method_novelty}a-b). In contrast, 90.4\% introduce new methods that do not appear in the seed papers. These results suggest that when AI-generated ideas differ from prior work, the differences arise predominantly from modifying or recombining methods rather than identifying new scientific problems.

Interestingly, the way AI-generated ideas build on prior work differs across scientific fields ({Figure~\ref{fig:task_method_novelty}c}). In engineering and the natural sciences, including Computer Science, Mathematics, Physics, Chemistry, Materials Science, and Engineering, AI-generated ideas almost always retain existing research questions, with most of the apparent difference arising from new methods or methodological recombination. By contrast, Sociology and Business exhibit substantially higher rates of new research questions and correspondingly lower rates of new methods.  This pattern is consistent with the nature of these fields, where emerging social and business phenomena frequently motivate new research problems, whereas innovation in engineering and the natural sciences more often takes the form of new methods for addressing established problems.

\section{Discussion and Implications}
This study suggests a more cautious interpretation of current AI research agents. Recent research introduces increasingly sophisticated agentic mechanisms into scientific ideation, including self-reflection, staged validation, role decomposition, and multi-agent deliberation~\citep{lu2026towards,baek2025researchagent,schmidgall2025agent}. These mechanisms can improve the coherence and plausibility of generated research proposals. 
However, our findings suggest that such capabilities do not necessarily translate into broader scientific exploration. Although these AI agents are specifically asked to propose novel,  high-impact, or unlike-prior-work ideas, AI-generated ideas remain substantially more concentrated than human-authored research, stay closer to the starting literature than later human follow-on work, align less with future research frontiers, and are estimated to have less scientific impact. 

This distinction matters because scientific discovery is not only about producing plausible ideas, but also about exploring the space of possible ideas. Human scientific progress often involves moving beyond established directions, exploring less familiar regions, and occasionally reframing the underlying research problem itself~\citep{foster2015tradition,uzzi2013atypical,fortunato2018science}. From this perspective, current AI research agents appear better suited to local elaboration than exploration. Our findings also suggest that increasingly sophisticated agentic AI frameworks and scaling LLM do not fundamentally resolve this limitation.

More broadly, our findings point toward a future challenge for AI-assisted scientific discovery. The central question may not only be whether AI systems can generate coherent scientific ideas, but whether they can help expand the range of scientific directions. As AI research agents become more deeply integrated into scientific workflows, designing agentic AI systems that broaden scientific exploration will become increasingly important.

\backmatter

\clearpage


\clearpage

\bibliography{arxiv}

\clearpage

\setcounter{section}{0}
\setcounter{subsection}{0}
\setcounter{figure}{0}
\setcounter{table}{0}
\setcounter{equation}{0}
\renewcommand{\thesection}{S\arabic{section}}
\renewcommand{\thesubsection}{\thesection.\arabic{subsection}}
\renewcommand{\thefigure}{S\arabic{figure}}
\renewcommand{\thetable}{S\arabic{table}}
\renewcommand{\theequation}{S\arabic{equation}}
\makeatletter
\@ifpackageloaded{hyperref}{
  \renewcommand{\theHsection}{supp.\arabic{section}}
  \renewcommand{\theHsubsection}{supp.\arabic{section}.\arabic{subsection}}
  \renewcommand{\theHfigure}{supp.\arabic{figure}}
  \renewcommand{\theHtable}{supp.\arabic{table}}
  \renewcommand{\theHequation}{supp.\arabic{equation}}
}{}
\makeatother

\startcontents[supplement]
\section*{Supplementary Information}

\setcounter{tocdepth}{2}
\printcontents[supplement]{}{1}{}
\clearpage

\section{Supplementary Methods}
\subsection{Data Sources and Research Area Construction}
\label{app:research_area}

\paragraph{Overview.}
We identify research areas using bibliographic coupling. The main idea is that papers studying related problems tend to cite the same prior work. We therefore represent each paper by the references it cites, transform these high-dimensional reference profiles into compact paper embeddings, and cluster papers with similar embeddings.

\paragraph{Corpus and metadata.}
We collect papers and citation links from the Semantic Scholar Academic Graph\footnote{https://www.semanticscholar.org/product/api}. The raw paper index contains titles, publication years, abstracts, citation counts, reference counts, venues, authors, external identifiers, and Semantic Scholar fields of study. The main ideation analysis uses 12 fields with sufficient title and context coherence after field-level quality checks, including Medicine, Biology, Engineering, Chemistry, Computer Science, Environmental Science, Materials Science, Physics, Mathematics, Economics, Business, and Sociology. 
We use papers published before 2020 to construct bibliographic-coupling representations. We use papers published between 2020-2025 in our main analysis. Table \ref{tab:corpus_stats} summarizes the Semantic Scholar corpus used in this study.

\begin{table}
\centering
\caption{Summary of the Semantic Scholar corpus and sampled ideation inputs used in the main analysis.}
\label{tab:corpus_stats}
\begin{tabular}{lr}
\toprule
Quantity & Value \\
\midrule
Semantic Scholar papers retained in paper index & 149,405,218 \\
Papers used for context construction & 75,019,333 \\
Fields used in main ideation analysis & 12 \\
Study years & 2020--2025 \\
Selected research areas & 155 \\
Seed-paper sets & 11,520 \\
\bottomrule
\end{tabular}
\end{table}

\paragraph{Bibliographic-coupling for paper representation.}
We construct research areas separately within each broad field. For a given field, let $\mathcal{P}$ denote its set of papers and $\mathcal{R}$ the set of corpus papers they cite.  

We represent each paper by its references, following the logic of bibliographic coupling~\cite{kessler1963bibliographic}: papers that cite overlapping prior literature tend to address related problems.
Under a matrix formulation, if $\mathbf{B} \in \{0,1\}^{|\mathcal{P}| \times |\mathcal{R}|}$ denotes the paper--reference matrix, then the standard bibliographic-coupling matrix is given by
\[
\mathbf{C} = \mathbf{B}\mathbf{B}^{\top},
\]
where each entry counts the number of references jointly cited by two papers.
The raw coupling matrix $\mathbf{C}$ captures shared citation structure, but it treats all references equally. In practice, some highly cited papers (e.g., widely used optimization methods or foundation architectures) are cited across many unrelated research areas and therefore provide relatively weak evidence of topical similarity.

Therefore, for each field, we construct paper citation embeddings directly from the paper--reference matrix $\mathbf{B}$. To reduce the influence of broadly cited references, we weight each reference column by an inverse-document-frequency term, so that references cited by many papers receive lower weight and more field-specific references receive higher weight. The weighted rows are L2-normalized, projected to $d=128$ dimensions using truncated SVD, and L2-normalized again to produce the bibliographic-coupling embeddings used for subsequent research area identification.

\paragraph{Research area identification.}
Research areas are identified by clustering the papers published in 2020--2025. Within each field, MiniBatchKMeans~\cite{sculley2010web} is applied to the embeddings to obtain candidate citation-defined areas, and papers are assigned to their nearest cluster centroid. We use MiniBatchKMeans because the corpus contains millions of papers per field, requiring a scalable clustering method that can be applied consistently across all analyzed fields.  We retain only active areas that contain papers in every study year from 2020 through 2025. This longitudinal filter yields 11,520 seed-paper sets that cover 155 distinct research areas across the 12 analyzed fields. Table~\ref{tab:core_research_areas} summarizes the number and examples of selected research areas in each field.

\begin{table}
\centering
\caption{Identified research areas by field. }
\label{tab:core_research_areas}
\scriptsize
\setlength{\tabcolsep}{3pt}
\begin{tabular}{p{3cm}rp{8cm}}
\toprule
Field & Areas & Representative selected areas \\
\midrule
Biology & 10 & Arabidopsis genetics; stem-cell differentiation; oocyte maturation; stem-cell culture; evolutionary cooperation; microbial community assembly \\
Business & 19 & Supply chains; insurance claims; supplier selection; digital platforms; live streaming; COVID-19 business disruption; financial reporting \\
Chemistry & 13 & DNA origami; gold nanoparticles; electronic structure; ion-mobility lipidomics; lipid membranes; molecular spectroscopy; polymer materials \\
Computer Science & 9 & Cognitive radio networks; error-correcting codes; image fusion; educational AI systems; machine-learning applications; networked agents; statistical evaluation \\
Economics & 18 & Synthetic control; time-series forecasting; propensity-score methods; digital currency; instrumental variables; treatment effects; financial risk \\
Engineering & 12 & Road traffic systems; ant-colony optimization; satellite communications; optogenetic control; power-system stability; structural monitoring; wireless networks \\
Environmental Science & 11 & Bayesian ecological modeling; parasite communities; persistent pollutants; heavy-metal contamination; wastewater treatment; microbial ecology; forest governance \\
Materials Science & 14 & Monolayer graphene; gold nanoparticles; first-principles materials modeling; lithium-ion batteries; iron oxide; nanocomposites; thin films \\
Mathematics & 10 & Partially hyperbolic dynamics; mapping class groups; least-squares estimation; Banach spaces; simple groups; partial differential equations; stochastic processes \\
Medicine & 9 & Metabolic syndrome; physical activity; pancreatic cancer; infectious disease; parasites; body-mass index; biomarkers; clinical risk factors \\
Physics & 15 & Quantum simulation; nonlinear wave equations; atomic gravimetry; solar coronal mass ejections; photonic crystals; finite-element methods; heavy-ion collisions \\
Sociology & 15 & Socioeconomic status; family planning; collective efficacy; aging; older adults; Indigenous communities; social inequality; public health behavior \\
\midrule
Total & 155 & \\
\bottomrule
\end{tabular}
\end{table}



\subsection{Generating Scientific Ideas with AI Agents}
\label{app:generation_scale}
This appendix provides additional implementation details for the ideation experiments described in the main text.  

\paragraph{Seed-paper contexts.}
We sample 11,520 seed-paper sets across the research areas, drawing 8 research areas per field per year and 20 anchor papers per area across the 2020--2025 study years. Each set contains five papers: one anchor paper and four related papers from the same research area. Related papers are selected from papers no later than the anchor year. All agent frameworks receive the same five-paper context for a given run, and the context contains only paper titles and abstracts.

\paragraph{AI Research Agent framework.}
We evaluate five AI research agent frameworks. The Zero-shot agent receives the five-paper context once and directly generates one research idea. AIScientist receives the same context, generates an initial idea, and then refines it through self-reflection rounds. In this study, we evaluate only this ideation stage of AIScientist~\cite{lu2026towards}, and we do not run its later experiment-execution tree-search stage. ResearchAgent~\cite{baek2025researchagent} uses the same five-paper context to propose a research problem, develop a method, and design experiments, with validator agents scoring intermediate outputs before final selection. AgentLaboratory~\cite{schmidgall2025agent} receives the same literature context as a short literature review and uses a dialogue between role-based agents, including a postdoc agent and a PhD-student agent, to formulate a final research plan. Co-Scientist~\cite{gottweis2026coscientist} is implemented based on the published description of the framework: multiple agents generate, reflect on, rank, evolve, and synthesize competing hypotheses before a final meta-review.

\paragraph{Prompts and tools.}
This section documents the prompts and tools used to generate ideas. Each system receives the same five paper titles and abstracts from the paper corpus available before time \(t\), but the systems differ in how they turn that evidence into a proposal. Zero-shot prompting uses one direct JSON prompt. AIScientist uses iterative ideation and reflection with a literature-search command. ResearchAgent uses separate problem, method, and experiment stages. AgentLaboratory uses a dialogue between role-based agents to produce a final plan.  Co-Scientist uses supervisor parsing, literature search, hypothesis generation, reflection, tournament ranking, evolution, and meta-review stages.

The placeholder \texttt{\{five\_paper\_literature\_context\}} in the prompt listings denotes the rendered five-paper context used at runtime. It is a numbered list containing only the title and abstract of each seed paper. For the zero-shot baseline, this rendered block is inserted directly at the beginning of the user message. For AIScientist, the same rendered block is passed as \texttt{\{workshop\_description\}} in the initial idea-generation prompt. For ResearchAgent, the first seed paper is passed through the original implementation's main-paper field, the remaining four seed papers are passed as references, and the entity list is left empty; our wrapper also keeps the full five-paper list for the simplified fallback prompts. For AgentLaboratory, the five papers are converted into the condensed literature-review field shown to the postdoc and PhD-student agents. For Co-Scientist, the five papers are passed to the supervisor stage as the initial research goal and literature context. Later stages use the parsed research problem, retrieved literature, and intermediate hypotheses. Other braced fields in the listings are filled with intermediate outputs from earlier stages of the same agent run.

\paragraph{Execution protocol.}
Each generation run starts from one seed-paper set, one agent framework, and one LLM. The zero-shot baseline makes a single model call and returns the JSON idea from that call. AIScientist first receives the five-paper context in the idea-generation prompt, then runs for five ideation/reflection rounds; at each round the model either issues a SearchSemanticScholar action or returns a FinalizeIdea action. The search action is served by a local literature-search wrapper over papers available before time \(t\), so the agent never observes follow-on papers from the evaluation period. ResearchAgent runs three stages in order: problem identification, method development, and experiment design. In the full ResearchAgent path, each stage is generated and validated for two iterations, and the highest-scoring candidate according to the validator is passed to the next stage. AgentLaboratory starts with the postdoc agent, alternates between postdoc and PhD-student turns for up to eight plan-formulation steps, and treats the postdoc's PLAN command as the final idea. Co-Scientist first generates competing hypotheses from the shared literature context, reflects on each hypothesis, ranks the candidates, and then writes a final synthesized proposal through a meta-review stage. If an agent does not produce the required final structured output, the run is marked invalid and excluded by the validity filter below.

\paragraph{Prompt listings.}
The boxes reproduce the prompt text used in the generation runs. For systems with multiple stages, the boxes are ordered in the same sequence as the generation process.

\noindent\emph{Zero-shot baseline.}
\begin{promptbox}
System:
You are an experienced researcher. Given a research topic and a set of relevant papers, propose exactly one novel, feasible research idea. Output requirements: emit ONLY a single JSON object that matches the schema given by the user. Begin your answer with '{' and end with '}'.

User:
{five_paper_literature_context}
---

Propose ONE novel research idea grounded in the literature above.
Reply with a single JSON object matching this schema:
{
  "Name": "<short snake_case identifier>",
  "Title": "<full paper title>",
  "Short Hypothesis": "<one sentence core claim>",
  "Related Work": "<how this differs from the provided papers>",
  "Abstract": "<~150 word abstract>",
  "Experiments": "<key experiments needed to validate the idea>",
  "Risk Factors and Limitations": "<main risks or limitations>"
}
\end{promptbox}

\noindent\emph{AIScientist.}
We use the original AIScientist ideation setup, which consists of a system prompt, an idea-generation user prompt, and a reflection user prompt. The agent runs for five rounds of ideation and reflection. It may call \texttt{SearchSemanticScholar} before finalizing an idea with \texttt{FinalizeIdea}. The prompt names this literature-search command as \texttt{SearchSemanticScholar}; in our runs, the command returns papers from our local paper corpus restricted to papers available before time \(t\).

\noindent\emph{AIScientist system prompt.}
\begin{promptbox}
You are an experienced AI researcher who aims to propose high-impact research ideas resembling exciting grant proposals. Feel free to propose any novel ideas or experiments; make sure they are novel. Be very creative and think out of the box. Each proposal should stem from a simple and elegant question, observation, or hypothesis about the topic. For example, they could involve very interesting and simple interventions or investigations that explore new possibilities or challenge existing assumptions. Clearly clarify how the proposal distinguishes from the existing literature.

Ensure that the proposal does not require resources beyond what an academic lab could afford. These proposals should lead to papers that are publishable at top conferences.

You have access to the following tools:

- **SearchSemanticScholar**: Search for relevant literature using Semantic Scholar. Provide a search query to find relevant papers.

- **FinalizeIdea**: Finalize your idea by providing the idea details.

The IDEA JSON should include the following fields:
- "Name": A short descriptor of the idea. Lowercase, no spaces, underscores allowed.
- "Title": A catchy and informative title for the proposal.
- "Short Hypothesis": A concise statement of the main hypothesis or research question. Clarify the need for this specific direction, ensure this is the best setting to investigate this idea, and there are not obvious other simpler ways to answer the question.
- "Related Work": A brief discussion of the most relevant related work and how the proposal clearly distinguishes from it, and is not a trivial extension.
- "Abstract": An abstract that summarizes the proposal in conference format (approximately 250 words).
- "Experiments": A list of experiments that would be conducted to validate the proposal. Ensure these are simple and feasible. Be specific in exactly how you would test the hypothesis, and detail precise algorithmic changes. Include the evaluation metrics you would use.
- "Risk Factors and Limitations": A list of potential risks and limitations of the proposal.

Respond in the following format:

ACTION:
<The action to take, exactly one of "SearchSemanticScholar", "FinalizeIdea">

ARGUMENTS:
<If ACTION is "SearchSemanticScholar", provide the search query as {"query": "your search query"}. If ACTION is "FinalizeIdea", provide the idea details as {"idea": { ... }} with the IDEA JSON specified below.>

If you choose to finalize your idea, provide the IDEA JSON in the arguments:

IDEA JSON:
```json
{
  "idea": {
    "Name": "...",
    "Title": "...",
    "Short Hypothesis": "...",
    "Related Work": "...",
    "Abstract": "...",
    "Experiments": "...",
    "Risk Factors and Limitations": "..."
  }
}
```

Ensure the JSON is properly formatted for automatic parsing.

Note: You should perform at least one literature search before finalizing your idea to ensure it is well-informed by existing research.
\end{promptbox}

\noindent\emph{AIScientist idea-generation user prompt.}
\begin{promptbox}
{workshop_description}

Here are the proposals that you have already generated:

'''
{prev_ideas_string}
'''

Begin by generating an interestingly new high-level research proposal that differs from what you have previously proposed.
\end{promptbox}

\noindent\emph{AIScientist reflection user prompt.}
\begin{promptbox}
Round {current_round}/{num_reflections}.

In your thoughts, first carefully consider the quality, novelty, and feasibility of the proposal you just created.
Include any other factors that you think are important in evaluating the proposal.
Ensure the proposal is clear and concise, and the JSON is in the correct format.
Do not make things overly complicated.
In the next attempt, try to refine and improve your proposal.
Stick to the spirit of the original idea unless there are glaring issues.

If you have new information from tools, such as literature search results, incorporate them into your reflection and refine your proposal accordingly.

Results from your last action (if any):

{last_tool_results}
\end{promptbox}

\noindent\emph{ResearchAgent.}
ResearchAgent formulates an idea through three stages: problem identification, method development, and experiment design. Each stage uses a role-specific system prompt and a user prompt. The evaluated ResearchAgent idea is constructed from the generated problem, method, and experiment plan. 

\noindent\emph{ProblemIdentifier system prompt.}
\begin{promptbox}
You are an AI assistant whose primary goal is to identify promising, new, and key scientific problems based on existing scientific literature, in order to aid researchers in discovering novel and significant research opportunities that can advance the field.
\end{promptbox}

\noindent\emph{ProblemIdentifier user prompt.}
\begin{promptbox}
You are going to generate a research problem that should be original, clear, feasible, relevant, and significant to its field. This will be based on the title and abstract of the target paper, those of {n_references} related papers in the existing literature, and {n_entities} entities potentially connected to the research area.

Understanding of the target paper, related papers, and entities is essential:
- The target paper is the primary research study you aim to enhance or build upon through future research, serving as the central source and focus for identifying and developing the specific research problem.
- The related papers are studies that have cited the target paper, indicating their direct relevance and connection to the primary research topic you are focusing on, and providing additional context and insights that are essential for understanding and expanding upon the target paper.
- The entities can include topics, keywords, individuals, events, or any subjects with possible direct or indirect connections to the target paper or the related studies, serving as auxiliary sources of inspiration or information that may be instrumental in formulating the research problem.

Your approach should be systematic:
- Start by thoroughly reading the title and abstract of the target paper to understand its core focus.
- Next, proceed to read the titles and abstracts of the related papers to gain a broader perspective and insights relevant to the primary research topic.
- Finally, explore the entities to further broaden your perspective, drawing upon a diverse pool of inspiration and information, while keeping in mind that not all may be relevant.

I am going to provide the target paper, related papers, and entities, as follows:

{target_paper_title_and_abstract}
{related_paper_titles_and_abstracts}
{entities}

With the provided target paper, related papers, and entities, your objective now is to formulate a research problem that not only builds upon these existing studies but also strives to be original, clear, feasible, relevant, and significant. Before crafting the research problem, revisit the title and abstract of the target paper, to ensure it remains the focal point of your research problem identification process.

{target_paper_title_and_abstract}
Then, following your review of the above content, please proceed to generate one research problem with the rationale, in the format of
Problem:
Rationale:
\end{promptbox}

\noindent\emph{MethodDeveloper system prompt.}
\begin{promptbox}
You are an AI assistant whose primary goal is to propose innovative, rigorous, and valid methodologies to solve newly identified scientific problems derived from existing scientific literature, in order to empower researchers to pioneer groundbreaking solutions that catalyze breakthroughs in their fields.
\end{promptbox}

\noindent\emph{MethodDeveloper user prompt.}
\begin{promptbox}
You are going to propose a scientific method to address a specific research problem. Your method should be clear, innovative, rigorous, valid, and generalizable. This will be based on a deep understanding of the research problem, its rationale, existing studies, and various entities.

Understanding of the research problem, existing studies, and entities is essential:
- The research problem has been formulated based on an in-depth review of existing studies and a potential exploration of relevant entities, which should be the cornerstone of your method development.
- The existing studies refer to the target paper that has been pivotal in identifying the problem, as well as the related papers that have been additionally referenced in the problem discovery phase, all serving as foundational material for developing the method.
- The entities can include topics, keywords, individuals, events, or any subjects with possible direct or indirect connections to the existing studies, serving as auxiliary sources of inspiration or information that may be instrumental in method development.

Your approach should be systematic:
- Start by thoroughly reading the research problem and its rationale, to understand your primary focus.
- Next, proceed to review the titles and abstracts of existing studies, to gain a broader perspective and insights relevant to the primary research topic.
- Finally, explore the entities to further broaden your perspective, drawing upon a diverse pool of inspiration and information, while keeping in mind that not all may be relevant.

I am going to provide the research problem, existing studies (target paper & related papers), and entities, as follows:

Research problem: {problem}
Rationale: {problem_rationale}

{target_paper_title_and_abstract}
{related_paper_titles_and_abstracts}
{entities}

With the provided research problem, existing studies, and entities, your objective now is to formulate a method that not only leverages these resources but also strives to be clear, innovative, rigorous, valid, and generalizable. Before crafting the method, revisit the research problem, to ensure it remains the focal point of your method development process.

Research problem: {problem}
Rationale: {problem_rationale}

Then, following your review of the above content, please proceed to propose your method with its rationale, in the format of
Method:
Rationale:
\end{promptbox}

\noindent\emph{ExperimentDesigner system prompt.}
\begin{promptbox}
You are an AI assistant whose primary goal is to design robust, feasible, and impactful experiments based on identified scientific problems and proposed methodologies from existing scientific literature, in order to enable researchers to systematically test hypotheses and validate groundbreaking discoveries that can transform their respective fields.
\end{promptbox}

\noindent\emph{ExperimentDesigner user prompt.}
\begin{promptbox}
You are going to design an experiment, aimed at validating a proposed method to address a specific research problem. Your experiment design should be clear, robust, reproducible, valid, and feasible. This will be based on a deep understanding of the research problem, scientific method, existing studies, and various entities.

Understanding of the research problem, scientific method, existing studies, and entities is essential:
- The research problem has been formulated based on an in-depth review of existing studies and a potential exploration of relevant entities.
- The scientific method has been proposed to tackle the research problem, which has been informed by insights gained from existing studies and relevant entities.
- The existing studies refer to the target paper that has been pivotal in identifying the problem and method, as well as the related papers that have been additionally referenced in the discovery phase of the problem and method, all serving as foundational material for designing the experiment.
- The entities can include topics, keywords, individuals, events, or any subjects with possible direct or indirect connections to the existing studies, serving as auxiliary sources of inspiration or information that may be instrumental in your experiment design.

Your approach should be systematic:
- Start by thoroughly reading the research problem and its rationale followed by the proposed method and its rationale, to pinpoint your primary focus.
- Next, proceed to review the titles and abstracts of existing studies, to gain a broader perspective and insights relevant to the primary research topic.
- Finally, explore the entities to further broaden your perspective, drawing upon a diverse pool of inspiration and information, while keeping in mind that not all may be relevant.

I am going to provide the research problem, scientific method, existing studies (target paper & related papers), and entities, as follows:

Research problem: {problem}
Rationale: {problem_rationale}

Scientific method: {method}
Rationale: {method_rationale}

{target_paper_title_and_abstract}
{related_paper_titles_and_abstracts}
{entities}

With the provided research problem, scientific method, existing studies, and entities, your objective now is to design an experiment that not only leverages these resources but also strives to be clear, robust, reproducible, valid, and feasible. Before crafting the experiment design, revisit the research problem and proposed method, to ensure they remain at the center of your experiment design process.

Research problem: {problem}
Rationale: {problem_rationale}

Scientific method: {method}
Rationale: {method_rationale}

Then, following your review of the above content, please proceed to outline your experiment with its rationale, in the format of
Experiment:
Rationale:
\end{promptbox}

\noindent\emph{Validator system prompts.}
\begin{promptbox}
ProblemValidator: You are an AI assistant whose primary goal is to assess the quality and validity of scientific problems across diverse dimensions, in order to aid researchers in refining their problems based on your evaluations and feedback, thereby enhancing the impact and reach of their work.
MethodValidator: You are an AI assistant whose primary goal is to assess the quality and soundness of scientific methods across diverse dimensions, in order to aid researchers in refining their methods based on your evaluations and feedback, thereby enhancing the impact and reach of their work.
ExperimentValidator: You are an AI assistant whose primary goal is to meticulously evaluate the experimental designs of scientific papers across diverse dimensions, in order to aid researchers in refining their experimental approaches based on your evaluations and feedback, thereby amplifying the quality and impact of their scientific contributions.
\end{promptbox}

\noindent\emph{AgentLaboratory.}
AgentLaboratory uses a dialogue between a Postdoc agent and a PhD-student agent during plan formulation. We evaluate the final \texttt{PLAN} command produced in this phase. In our setting, this plan-formulation phase uses the provided literature context and dialogue history.

\noindent\emph{Shared inference templates.}
\begin{promptbox}
You are {role_description}
Task instructions: {phase_prompt}
{command_descriptions}

{context}
~~~~~~~~~~
History: {history_str}
~~~~~~~~~~
Current Step #{step}, Phase: plan formulation
{complete_str}
[Objective] Your goal is to perform research on the following topic: {research_topic}
Feedback: {feedback}
Notes: {notes_str}
Your previous command was: {prev_comm}. Make sure your new output is very different.
Please produce a single command below:
\end{promptbox}

\noindent\emph{Postdoc role description and phase prompt.}
\begin{promptbox}
role_description:
a postdoctoral student at a top university.

phase_prompt:
You are directing a PhD student to help them come up with a good plan, and you interact with them through dialogue.
Your goal is to produce plans that would make good experiments for the given topic. You should aim for a very simple experiment that showcases your plan, not a complex one. You should integrate the provided literature review and come up with plans on how to expand and build on these works for the given topic. Your plans should provide a clear outline for how to achieve the task, including what machine learning models to use and implement, what types of datasets should be searched for and used to train the model, and the exact details of the experiment. Your idea should be very innovative and unlike anything seen before.
\end{promptbox}

\noindent\emph{Postdoc command descriptions.}
\begin{promptbox}
You can produce dialogue using the following command: ```DIALOGUE
dialogue here
```
 where dialogue here is the actual dialogue you will send and DIALOGUE is just the word DIALOGUE.
When you believe a good plan has been arrived at between you and the PhD student you can use the following command to end the dialogue and submit the plan ```PLAN
plan here
```
 where plan here is the actual plan to be transmitted and PLAN is just the word PLAN. Plan here should provide a clear outline for how to achieve the task, including what machine learning models to use and implement, what types of datasets should be searched for and used to train the model, and the exact details of the experiment.
You can only use a SINGLE command per inference turn. Do not use more than one command per inference. If you use multiple commands, then only one of them will be executed, NOT BOTH.
Make sure not to produce too much dialogue and to submit an plan in reasonable time.When performing a command, make sure to include the three ticks (```) at the top and bottom ```COMMAND
text
``` where COMMAND is the specific command you want to run (e.g. PLAN, DIALOGUE).
\end{promptbox}

\noindent\emph{PhD-student role description and phase prompt.}
\begin{promptbox}
role_description:
a PhD student at a top university.

phase_prompt:
You are a PhD student being directed by a postdoc who will help you come up with a good plan, and you interact with them through dialogue.
Your goal is to produce plans that would make good experiments for the given topic. You should aim for a very simple experiment that showcases your plan, not a complex one. You should integrate the provided literature review and come up with plans on how to expand and build on these works for the given topic. Your plans should provide a clear outline for how to achieve the task, including what machine learning models to use and implement, what types of datasets should be searched for and used to train the model, and the exact details of the experiment. Your idea should be very innovative and unlike anything seen before.
\end{promptbox}

\noindent\emph{PhD-student command descriptions.}
\begin{promptbox}
You can produce dialogue using the following command: ```DIALOGUE
dialogue here
```
 where 'dialogue here' is the actual dialogue you will send and DIALOGUE is just the word DIALOGUE.
You can only use a single command per inference turn. Do not use more than one command per inference. If you use multiple commands, then only one of them will be executed, not both.
When performing a command, make sure to include the three ticks (```) at the top and bottom ```COMMAND
text
``` where COMMAND is the specific command you want to run (e.g. DIALOGUE).
\end{promptbox}

\noindent\emph{Co-Scientist.}
Co-Scientist follows the published generate--debate--evolve workflow. A supervisor agent parses the research goal and search queries; generation agents propose hypotheses; reflection agents critique them; ranking agents compare hypotheses by novelty, significance, feasibility, and testability; evolution agents revise top hypotheses; and a meta-review agent synthesizes the final proposal.

\noindent\emph{Co-Scientist supervisor prompt.}
\begin{promptbox}
System:
You are the Supervisor agent in a Co-Scientist system. Parse the user's research goal into a compact research plan and search queries. Return only JSON.

User:
Research goal and literature context:
{research_goal_and_literature_context}

Return JSON with keys: research_problem, constraints, search_queries. search_queries must be 3 to 5 short literature search queries.
\end{promptbox}

\noindent\emph{Co-Scientist generation prompt.}
\begin{promptbox}
System:
You are the Generation agent from a Co-Scientist system. Create specific, testable, non-obvious research hypotheses grounded in the literature. Return only JSON.

User:
Research problem:
{research_problem}

Retrieved literature:
{retrieved_literature}

Generation strategy: {generation_strategy}.
Produce {count} distinct hypotheses. For each, include title, hypothesis, rationale, experiments, expected_evidence, limitations, and novelty_claim.
Return JSON: {"hypotheses": [ ... ]}.
\end{promptbox}

\noindent\emph{Co-Scientist reflection prompt.}
\begin{promptbox}
System:
You are the Reflection agent in a Co-Scientist system. Review hypotheses for novelty, correctness, feasibility, testability, and hidden assumptions. Return only JSON.

User:
Literature context:
{retrieved_literature}

Hypothesis to review:
{hypothesis}

Return JSON with keys: novelty_score, correctness_score, feasibility_score, testability_score, key_weaknesses, suggested_revision, verdict.
\end{promptbox}

\noindent\emph{Co-Scientist ranking prompt.}
\begin{promptbox}
System:
You are the Ranking agent in a Co-Scientist tournament. Compare two hypotheses by novelty, significance, feasibility, and testability. Return only JSON.

User:
Compare these two hypotheses. Simulate a concise scientific debate, then choose a winner.

Hypothesis A:
{hypothesis_a}

Hypothesis B:
{hypothesis_b}

Return JSON with keys: debate_summary, winner (A/B/tie), rationale.
\end{promptbox}

\noindent\emph{Co-Scientist evolution prompt.}
\begin{promptbox}
System:
You are the Evolution agent in a Co-Scientist system. Improve top hypotheses by combining strengths, improving feasibility, or proposing an out-of-the-box variant. Return only JSON.

User:
Retrieved literature:
{retrieved_literature}

Top-ranked hypotheses:
{top_ranked_hypotheses}

Generate {count} evolved hypotheses. Return JSON: {"hypotheses": [ ... ]}.
\end{promptbox}

\noindent\emph{Co-Scientist meta-review prompt.}
\begin{promptbox}
System:
You are the Meta-review agent in a Co-Scientist system. Synthesize the hypothesis tournament and reviews into one final research proposal. Return only JSON.

User:
Ranked hypotheses and reviews:
{ranked_hypotheses_and_reviews}

Return JSON matching this schema: {"Name": str, "Title": str, "Short Hypothesis": str, "Related Work": str, "Abstract": str, "Experiments": str, "Risk Factors and Limitations": str, "ranked_hypotheses": list}.
\end{promptbox}

For reproducibility, we store the rendered prompt text, raw model responses, parsed actions, and intermediate stage outputs for each run.

\paragraph{Output standardization.}
The evaluated AI research-agent frameworks produce outputs in substantially different formats, including structured JSON ideas, staged research proposals, markdown plans, and multi-agent dialogue traces. Before analysis, we first convert the output of each framework into a standardized generated-idea document. For Zero-shot, AIScientist, and Co-Scientist, we use the generated title (or name), hypothesis, and abstract-like proposal text. For ResearchAgent, we use the proposed research problem and method. For Agent Laboratory, we use the main proposal sections of the final research plan. This standardization produces a unified textual representation of every generated idea.

Although the generated ideas have been standardized into a common textual format, they remain heterogeneous with respect to human-authored papers, which are represented by titles and abstracts. To enable a unified comparison, we further convert both AI-generated ideas and human-authored papers into a common scholarly annotation schema following recent scientific-article key-insight extraction approaches~\citep{song2025scientific}. Specifically, we prompt Gemma-4-31B-IT to extract the Aim, Motivation, Research Question, Technical Method, and Scholarly Keywords for every document. The extracted research questions and technical methods are concatenated and encoded into the shared semantic embedding space used throughout the paper.

The extraction prompt is shown below:
\begin{promptbox}
System:
You are a careful scholarly annotator. Given a research manuscript, write a concise scholarly analysis covering Aim, Motivation, Questions addressed, Method, Evaluation metrics, Findings, Contributions, Limitations, and Future work. Finally extract 5-12 concise scholarly keywords grounded in that analysis. Return exactly one valid JSON object, with no markdown, no prose outside the JSON, and no missing JSON fields.

User:
Document kind: {paper_or_generated_idea}
Document id: {document_id}

Please proceed to conduct a scholarly analysis of the provided research manuscript. Your analysis should encapsulate the core components of the study as delineated in the enumeration below:
Aim: What is the aim of the study?
Motivation: What is the motivation of the study?
Questions addressed: What question does this study address?
Methods: What methods does the study use to solve the question?
Evaluation metrics: What evaluation metrics are used in this study?
Findings: What does the study find?
Contributions: What are the contributions of this study?
Limitations: What are the limitations of this study?
Future work: What is the future work of this study?

Subsequently, organize the distilled information into a structured JSON format, omitting any supplementary explanations.

Return exactly one JSON object with this structure:

{{
  "analysis": {{
    "Aim": "...",
    "Motivation": "...",
    "Questions addressed": "...",
    "Method": "...",
    "Evaluation metrics": "...",
    "Findings": "...",
    "Contributions": "...",
    "Limitations": "...",
    "Future work": "..."
  }},
  "keywords": ["...", "..."]
}}

Rules:
- Output JSON only; do not wrap it in markdown fences.
- Include every analysis field exactly as shown above.
- keywords must be a non-empty list of 5-12 short scholarly noun phrases.
- If a field is uncertain, write a brief best-effort value rather than omitting it.

Research manuscript:
{document_text}

\end{promptbox}

\paragraph{Idea Generation and Validity Filtering.}
For the 11,520 seed-paper sets, we evaluate four open-source LLMs under five AI agent frameworks, yielding 230,400 generation runs. Due to budget constraints, experiments using the proprietary GPT-5.4 model are conducted only on randomly sampled seed-paper sets from 2022, resulting in an additional 2,400 generation runs. In total, the study includes 232,800 AI idea-generation runs.

We next apply validity filtering to remove unsuccessful generations. A generation run is considered valid if the agent completes successfully and produces a non-empty structured output from which a research idea can be extracted. Runs that fail to complete, return unparsable structured outputs, or produce an empty final proposal are excluded. This filtering yields 219,655 valid AI-generated ideas from 232,800 generation runs, corresponding to a validity rate of 94.4\%.

Table~\ref{tab:generation_scale_2} summarizes the numbers of generation runs and valid ideas by publication year, AI agent framework, and LLM. Table~\ref{tab:idea_examples} presents representative AI-generated ideas produced by Gemma-4-31B-IT under different agent frameworks using the same seed-paper set.

\begin{table}[htbp]
\centering
\caption{Breakdown of the ideation design across study years, agent frameworks, and LLMs.}
\label{tab:generation_scale_2}
\begin{tabular}{lrr}
\toprule
\textbf{Dimension} & \textbf{Categories} & \textbf{Generation runs} \\
\midrule
\multicolumn{3}{l}{\textit{Overall}} \\
\quad Total open-weight runs & $5 \times 4 \times 11{,}520$ & 230,400 \\
\quad GPT-5.4 subset runs & 2022 representative subset & 2,400 \\
\quad Total generation runs & Open-weight + GPT-5.4  & \textbf{232,800} \\
\quad Total AI-generated Ideas & Valid output & \textbf{219,655} \\
\midrule
\multicolumn{3}{l}{\textit{By year (1,920 seed sets $\times$ 5 agents $\times$ 4 LLMs)}} \\
\quad 2020 & 1,920 seed sets & 38,400 \\
\quad 2021 & 1,920 seed sets & 38,400 \\
\quad 2022 & 1,920 seed sets & 38,400 \\
\quad 2023 & 1,920 seed sets & 38,400 \\
\quad 2024 & 1,920 seed sets & 38,400 \\
\quad 2025 & 1,920 seed sets & 38,400 \\
\midrule
\multicolumn{3}{l}{\textit{By agent (11,520 seed sets $\times$ 4 LLMs)}} \\
\quad Agent Laboratory & 1 agent & 46,080 \\
\quad AIScientist & 1 agent & 46,080 \\
\quad ResearchAgent & 1 agent & 46,080 \\
\quad Co-Scientist & 1 agent & 46,080 \\
\quad Zero-shot & 1 agent & 46,080 \\
\midrule
\multicolumn{3}{l}{\textit{By LLM (11,520 seed sets $\times$ 5 agents)}} \\
\quad Gemma-4-31B-IT & Open-weight  & 57,600 \\
\quad Llama-3.1-8B & Open-weight & 57,600 \\
\quad Hermes-4-14B & Open-weight & 57,600 \\
\quad Qwen3-35B-A3B & Open-weight & 57,600 \\
\quad GPT-5.4 & Proprietary & 2,400 \\
\bottomrule
\end{tabular}
\end{table}

\begin{table}
\centering
\caption{Examples of AI-generated ideas. Gemma-4-31B-IT is used as the LLM.}
\label{tab:idea_examples}
\small
\setlength{\tabcolsep}{5pt}
\renewcommand{\arraystretch}{1.22}
\begin{tabular}{@{}p{2.4cm}p{10.7cm}@{}}
\toprule
\textbf{Agent} & \textbf{Generated idea} \\
\midrule
\multicolumn{2}{@{}p{13.1cm}@{}}{\textbf{2025 run 1: long-range graph representation learning.}
\emph{Learning Long Range Dependencies on Graphs Via Random Walks~\cite{chen2025learning}; Polynormer~\cite{deng2024polynormer}; Classic GNNs Are Strong Baselines~\cite{luo2024classic}; Diffusing to the Top~\cite{lin2025diffusing}; Predicting Global Label Relationship Matrix under Heterophily~\cite{liang2023predicting}.}} \\
\addlinespace[2pt]
Zero-shot &
\textbf{PolyWalker.}
Replaces random-walk sequence encoders with polynomial-expressive linear attention to capture long-range graph dependencies efficiently. \\
AIScientist &
\textbf{GraphMamba.}
Uses selective state-space models over random-walk sequences to scale long-range graph dependency modeling with linear memory. \\
ResearchAgent &
\textbf{PolyWalk-LR.}
Combines local message passing, random-walk embeddings, polynomial attention, and a low-rank global label-relation matrix. \\
Agent Lab. &
\textbf{PolyDiff-GNN.}
Adds a plug-and-play polynomial diffusion layer to classic GNNs to model multi-hop dependencies without graph-transformer cost. \\
Co-Scientist &
\textbf{AdaptiveWalk-GNN.}
Learns node-adaptive gates over random-walk sequence features, polynomial diffusion scales, and local message passing, with label-relation regularization for heterophilous graphs. \\
\addlinespace[4pt]
\midrule
\multicolumn{2}{@{}p{13.1cm}@{}}{\textbf{2025 run 2: causal visual reasoning and tool-use agents.}
\emph{NarrativeBridge~\cite{nadeem2025narrativebridge}; CS-Bench~\cite{song2025csbench}; TACT~\cite{caciularu2024tact}; VisMin~\cite{awal2024vismin}; GTA: A Benchmark for General Tool Agents~\cite{wang2024gta}.}} \\
\addlinespace[2pt]
Zero-shot &
\textbf{CausalVisTool.}
Combines causal-temporal video narratives with tool-use evaluation to test whether multimodal agents preserve causal consistency. \\
AIScientist &
\textbf{Counterfactual-VideoBench.}
Turns causal video understanding into a counterfactual intervention benchmark for testing whether VLMs rely on temporal correlations. \\
ResearchAgent &
\textbf{MATR.}
Links fine-grained visual perception, aggregative information extraction, and tool execution in a perceive--aggregate--execute pipeline. \\
Agent Lab. &
\textbf{CCTR.}
Evaluates causal tool reasoning by asking agents to connect dynamic visual events with the correct tool actions and counterfactual variants. \\
Co-Scientist &
\textbf{InterveneBench.}
Pairs minimally edited counterfactual videos with executable tool trajectories to test whether agents revise tool choices when causal events change while irrelevant visual content remains fixed. \\
\bottomrule
\end{tabular}
\end{table}

\newpage
\subsection{Definitions of Exploration Measures}
\label{app:metric_definitions}

This section formalizes the four measures introduced in the main text. Each AI-generated idea or human-authored paper is represented by its standardized text, from which we compute a text embedding and extract scholarly keywords. Below, $\mathbf{e}_x$ denotes the L2-normalized embedding of an idea or paper $x$. In the paper, we use Qwen3-Embedding-4B~\citep{zhang2025qwen3} as the text embedding model.

\textbf{Exploration breadth} measures how widely a set of ideas or papers spreads within the same research area. Let $X_a = \{x_1, \ldots, x_n\}$ contain either the AI-generated ideas from research area $a$ or human-authored papers from that area. Breadth is the mean pairwise cosine distance,
\[
\mathrm{Breadth}(X_a)
= \frac{2}{n(n-1)} \sum_{1 \le i < j \le n} \bigl(1 - \mathbf{e}_{x_i}^{\top}\mathbf{e}_{x_j}\bigr).
\]
The values reported in the main text average this quantity over research areas within each comparison group (pooled, by agent framework, or by LLM). 

\textbf{Exploration distance} measures how far an idea or paper moves from the seed literature. Each generation run $r$ starts from a set $S_r$ of five seed papers, summarized by the normalized centroid
\[
\mathbf{c}_r = \frac{\sum_{p \in S_r} \mathbf{e}_p}{\bigl\lVert \sum_{p \in S_r} \mathbf{e}_p \bigr\rVert_2}.
\]
AI-generated ideas from run $r$ and follow-on human papers citing at least one paper in $S_r$ are scored by their cosine distance to this centroid,
\[
\mathrm{Dist}(x; r) = 1 - \mathbf{e}_x^{\top}\mathbf{c}_r,
\]
so that ideas and papers stay close to the seed literature receive small distances.

\textbf{Frontier alignment} measures how well a group of ideas or papers covers the topics that become prominent in the following year. For field $f$ and seed year $t$, the next-year frontier $F_{f,t+1}$ is the set of the top 10\% most frequent scholarly keywords among human-authored papers published in field $f$ in year $t+1$, excluding the evaluated follow-on papers. For each comparison group $g$, the keywords extracted from its ideas or papers are pooled into a set $K_g$, and frontier alignment is the share of the frontier covered by this pool,
\[
\mathrm{Alignment}(g) = \frac{\lvert K_g \cap F_{f,t+1} \rvert}{\lvert F_{f,t+1} \rvert}.
\]

\textbf{Potential scientific impact} measures whether an idea or paper falls near historically influential parts of its research area. Each human-authored paper $p$ from research area $a$ and publication year $y$ receives a normalized citation score
\[
s_p = \log(1 + c_p) - \bar{\ell}_{a,y}^{(-p)},
\]
where $c_p$ is its citation count and $\bar{\ell}_{a,y}^{(-p)}$ is the leave-one-out mean of $\log(1 + c_q)$ over the other papers from the same area and year. The potential scientific impact of an AI-generated idea or follow-on human paper $x$ is the mean score of its $k = 20$ nearest human-authored papers in the same area,
\[
\mathrm{Impact}(x) = \frac{1}{\lvert N_k(x) \rvert} \sum_{p \in N_k(x)} s_p,
\]
where $N_k(x)$ contains the $k$ nearest papers by cosine similarity, restricted to papers published no later than the seed year.

\section{Supplementary Discussion}

\subsection{Results by Scientific Field}
\label{app:field_variation}
The main text reports average results across the 12 broad scientific fields. Here we present field-level results for all four measures. For each measure, the reported gap is defined as the AI-generated idea score minus the corresponding human-paper score. Negative values therefore indicate that AI-generated ideas have lower exploration breadth, shorter exploration distance, lower frontier alignment, or lower potential scientific impact than the corresponding human-authored papers. The detailed 
results are presented in Table \ref{tab:field_variation}.

\begin{table}[t]
\centering
\caption{\textbf{Field-level results.} Each gap is computed as the AI-generated idea score minus the corresponding human-paper score. Thus, negative values indicate that AI-generated ideas have lower exploration breadth, exploration distance, frontier alignment, or potential scientific impact than human-authored papers. Fields are ordered by the impact gap from largest to smallest. $^{**}P<0.001$, $^{*}P<0.01$.} 
\label{tab:field_variation}
\small
\setlength{\tabcolsep}{6pt}
\renewcommand{\arraystretch}{1.12}
\begin{tabular*}{\textwidth}{@{\extracolsep{\fill}}lrrrr@{}}
\toprule
Field & Breadth gap & Distance gap & Frontier gap & Impact gap \\
\midrule
Computer Science & -0.032$^{**}$ & -0.069$^{**}$ & -0.068$^{**}$ & -0.161$^{**}$ \\
Business & -0.041$^{**}$ & -0.059$^{**}$ & -0.084$^{**}$ & -0.143$^{**}$ \\
Sociology & -0.045$^{**}$ & -0.037$^{**}$ & -0.119$^{**}$ & -0.109$^{**}$ \\
Materials Science & -0.024$^{*}$ & -0.086$^{**}$ & -0.103$^{**}$ & -0.101$^{**}$ \\
Engineering & -0.047$^{**}$ & -0.079$^{**}$ & -0.066$^{**}$ & -0.089$^{**}$ \\
Chemistry & -0.027$^{**}$ & -0.072$^{**}$ & -0.066$^{**}$ & -0.065$^{**}$ \\
Environmental Science & -0.030$^{*}$ & -0.052$^{**}$ & -0.071$^{**}$ & -0.058$^{**}$ \\
Medicine & -0.025$^{*}$ & -0.040$^{**}$ & -0.141$^{**}$ & -0.050$^{**}$ \\
Economics & -0.041$^{**}$ & -0.083$^{**}$ & -0.039$^{**}$ & -0.037$^{*}$ \\
Physics & -0.023$^{**}$ & -0.056$^{**}$ & -0.050$^{**}$ & -0.032$^{*}$ \\
Biology & -0.028$^{**}$ & -0.088$^{**}$ & -0.052$^{**}$ & -0.028$^{*}$ \\
Mathematics & 0.008 & -0.045$^{**}$ & -0.101$^{**}$ & -0.006 \\
\bottomrule
\end{tabular*}
\end{table}

\subsection{Robustness Check: Centroid-Based Measure of Exploration Breadth}
\label{app:centroid_radius}
We measure exploration breadth using an alternative centroid-based approach. For each research area and each comparison group (i.e., AI-generated ideas produced by a specific agent framework, LLM, and publication year), we compute the normalized centroid of the semantic embeddings and measure the cosine distance between each AI-generated idea and its corresponding centroid. Larger distances indicate greater exploration breadth, whereas smaller distances indicate that papers or ideas are more tightly concentrated around the centroid of their research area.

Table~\ref{tab:centroid_radius} reports the resulting centroid-distance statistics across AI agent frameworks and LLMs. Consistent with the main pairwise-similarity analysis, AI-generated ideas remain closer to their area centroids than human-authored papers do. The pattern holds for each agent framework and each LLM, including Co-Scientist and GPT-5.4.

\begin{table}[htbp]
\centering
\caption{Centroid-based robustness check for exploration breadth. Distances are cosine distances to the normalized within-area centroid. Lower values indicate tighter concentration.}
\label{tab:centroid_radius}
\small
\begin{tabular}{lrr}
\toprule
Group & Mean dist. & Median dist. \\
\midrule
AI ideas & 0.340 & 0.335 \\
Human papers & 0.362 & 0.355 \\
\midrule
\multicolumn{3}{@{}c@{}}{\textbf{AI ideas by {agent framework}}} \\
\addlinespace[2pt]
Zero-shot & 0.319 & 0.314 \\
AIScientist & 0.320 & 0.315 \\
ResearchAgent & 0.338 & 0.329 \\
Agent Lab. & 0.319 & 0.315 \\
Co-Scientist & 0.334 & 0.326 \\
\midrule
\multicolumn{3}{@{}c@{}}{\textbf{AI ideas by {LLM}}} \\
\addlinespace[2pt]
Llama-3.1-8B & 0.319 & 0.314 \\
Hermes-4-14B & 0.325 & 0.318 \\
Gemma-4-31B-IT & 0.332 & 0.328 \\
Qwen3-35B-A3B & 0.340 & 0.335 \\
GPT-5.4 & 0.303 & 0.298 \\
\bottomrule
\end{tabular}
\end{table}

\subsection{Validation of the Potential Scientific Impact Measure}
\label{app:local_impact_discussion}
We estimate the potential scientific impact of AI-generated ideas using the average normalized citation score of their 20 nearest human-paper neighbors. Here we validate this neighborhood-based impact measure using a leave-one-out analysis on human-authored papers.
Specifically, for each target human-authored paper, we remove the target paper itself and compute the average normalized citation score of its 20 nearest prior human-paper neighbors from the same research area. We then examine whether this neighborhood-based impact score predicts the target paper's own normalized citation score.

Table~\ref{tab:local_impact_validation} shows that the neighborhood-based impact score is positively associated with the target paper's subsequent citation performance (Spearman $\rho=0.155$, Pearson $r=0.166$; both $P<0.001$). Thus, papers located in semantic neighborhoods with historically higher citation impact are themselves more likely to become highly cited. Although local semantic neighborhoods explain only part of the variation in citation outcomes, these results support the use of neighborhood citation statistics as a proxy for estimating the potential scientific impact of AI-generated ideas.

\begin{table}[t]
\centering
\caption{Validation of the impact score in human papers. Each target paper is scored using the mean citation score of its 20 nearest prior human-paper neighbors from the same research area. $^{**}P<0.001$.}
\label{tab:local_impact_validation}
\small
\setlength{\tabcolsep}{6pt}
\renewcommand{\arraystretch}{1.12}
\begin{tabular*}{0.62\textwidth}{@{\extracolsep{\fill}}lr@{}}
\toprule
Measure & Value \\
\midrule
Human papers evaluated & 16,500 \\
Mean prior neighbors & 19.9 \\
Spearman correlation & 0.155$^{**}$ \\
Pearson correlation & 0.166$^{**}$ \\
\bottomrule
\end{tabular*}
\end{table}

\subsection{Validation of Research Question and Method Annotation}\label{app:label_agreement}
The analysis in Section~4.6 relies on identifying the research question and technical methods of each AI-generated idea and determining whether they are already present in the corresponding five seed papers. To evaluate the reliability of this annotation procedure, we conduct an independent LLM-based validation following recent work on LLM annotation reliability~\citep{gilardi2023chatgpt}.

Specifically, three independent LLM annotators, including Qwen-30b, Llama-8B and Gemma-31B, are given the same AI-generated idea together with its corresponding five seed papers. Each annotator independently determines whether the generated idea introduces (i) a research question that is absent from the seed literature and (ii) a technical method that is absent from the seed literature. 
Table~\ref{tab:recombination_label_agreement} summarizes the agreement among the three annotators. Agreement is consistently high for both research question and research method judgments. All three annotators agree on 74.0\% of research-question labels and 77.6\% of technical-method labels. Pairwise agreement ranges from 80.8\% to 89.9\%, while Gwet's AC1 exceeds 0.77 for both tasks despite the class imbalance. These results indicate that determining whether an AI-generated idea introduces a new research question or a new method relative to the seed literature is a reliable annotation task.

\begin{table}[htbp]
\centering
\caption{Agreement among three LLM annotators for new-question and new-method-or-approach labels. Each annotator compares the generated idea with the same five seed papers. Values are computed over the common set of generated ideas with valid labels. Gwet's AC1 is reported because both binary labels are imbalanced~\cite{gwet2002kappa}.}
\label{tab:recombination_label_agreement}
\small
\setlength{\tabcolsep}{5pt}
\renewcommand{\arraystretch}{1.12}
\begin{tabular}{lrr}
\toprule
Agreement metric & Research question & Method \\
\midrule
All three annotators agree & 74.0\% & 77.6\% \\
Gwet's AC1 & 0.771 & 0.809 \\
Qwen 30B vs. Llama 8B & 82.7\% & 80.8\% \\
Qwen 30B vs. Gemma 31B & 82.9\% & 84.5\% \\
Llama 8B vs. Gemma 31B & 82.3\% & 89.9\% \\
\bottomrule
\end{tabular}
\end{table}

\clearpage

\end{document}